\documentclass[preprint,11pt]{article}

\usepackage{amssymb}
\usepackage[utf8]{inputenc}
\usepackage[numbers]{natbib}
\usepackage{multicol}
\usepackage{titlesec}
\usepackage[colorinlistoftodos]{todonotes}
\usepackage{caption}
\usepackage{subcaption}
\usepackage{url}
\newcommand{\etal}{{\em et~al.}}
\newcommand{\etc}{{\em etc.}}

\title{FRAMED: An AutoML Approach for Structural Performance Prediction of Bicycle Frames}

\author{
  Lyle Regenwetter, Colin Weaver Faez Ahmed\\
  Massachusetts Institute of Technology \\
  \texttt{\{regenwet, weaverc, faez\}@mit.edu} \\
}

\begin{document}
\maketitle
\begin{abstract}
This paper demonstrates how Automated Machine Learning (AutoML) methods can be used as effective surrogate models in engineering design problems. To do so, we consider the challenging problem of structurally-performant bicycle frame design and demonstrate across-the-board dominance by AutoML in regression and classification surrogate modeling tasks.
We also introduce FRAMED -- a parametric dataset of 4500 bicycle frames based on bicycles designed by practitioners and enthusiasts worldwide. Accompanying these frame designs, we provide ten structural performance values such as weight, displacements under load, and safety factors computed using finite element simulations for all the bicycle frame designs. We formulate two challenging test problems: a performance-prediction regression problem and a feasibility-prediction classification problem.  We then systematically search for optimal surrogate models using Bayesian hyperparameter tuning and neural architecture search. Finally, we show how a state-of-the-art AutoML method can be effective for both regression and classification problems. We demonstrate that the proposed AutoML models outperform the strongest gradient boosting and neural network surrogates identified through Bayesian optimization by an improved F1 score of 24\% for classification and reduced mean absolute error by 12.5\% for regression.
Our work introduces a dataset for bicycle design practitioners, provides two benchmark problems for surrogate modeling researchers, and demonstrates the advantages of AutoML in machine learning tasks. The dataset and code are provided at \url{http://decode.mit.edu/projects/framed/}.
\end{abstract}

\section{Introduction}
Data-driven methods have shown great promise in accelerating design tasks and enabling design automation across countless design domains. Data-driven approaches to design can tap into the immeasurable expertise captured within existing designs ranging from early-stage design concepts to rough prototypes to products on the market. Designers can leverage design information implicitly embedded in quality data to accelerate their own design process. In particular, tools like surrogate models trained on design data can help designers rapidly evaluate early-stage design concepts without the need for expensive and time-consuming simulation or physical experimentation. 

Two common surrogate modeling tasks are design performance prediction and validity evaluation. We formulate two corresponding benchmark problems related to performance-aware bicycle frame design. As one of the most ubiquitous consumer products in our modern society, the bicycle makes for a high-impact design problem. Making bicycles accessible to more people has numerous societal benefits, such as boosting public health~\cite{oja2011health}, mitigating traffic congestion~\cite{hamilton2018bicycle}, and reducing emissions~\cite{edenhofer2015climate}. These tantalizing prospects provide ample motivation to increase accessibility to bicycles and to improve their performance to raise appeal. With some estimates putting the number of privately owned bicycles at over 580 million~\cite{oke2015tracking}, even incremental improvements in bicycle design methodology would undoubtedly have an immense impact. We hope that data-driven surrogate models can accelerate the design process of customized bicycle frames, making them cheaper and more performant. 

Our data-driven approach to bicycle frame design and optimization delivers the following key contributions:
\begin{itemize}
    \item We introduce a dataset of 4500 bicycle frames adapted from bicycles designed by community members using the BikeCAD software. For each frame, we provide ten structural performance indicators evaluating the frame's performance under three load cases (in-plane, transverse, and eccentric loading). Indicators consist of seven deflections, two safety factors, and a weight value and are calculated through Finite Element Analysis.
    \item We validate our Finite Element Analysis framework through a mesh convergence study and verify the accuracy of our simulation results against physical testing of bicycle frames.
    \item We identify optimal surrogate models which predict the performance and feasibility of frames. Surrogates are selected using an AutoML framework which automates the selection of algorithms, architectures, hyperparameters, and instantiations for optimal model performance. AutoML models achieved a coefficient of determination of 0.605 in structural performance prediction and an F1 score of 0.915 in feasibility classification. 
    \item We validate our proposed AutoML framework against several common models (Neural Networks, XGBoost, etc.), which we optimize using Bayesian hyperparameter tuning. The proposed AutoML models attain the best coefficient of determination and mean absolute error among all methods tested in regression and the best F1 score, precision, recall, accuracy, and ROC AUC in classification.

\end{itemize}

\section{Background}
This paper explores the application of data-driven predictive models to bicycle frame design. In this section, we review existing literature on structural optimization of bicycle frames, emphasizing previous data-driven approaches. We then introduce key ideas and methods in supervised Machine Learning and Automated Machine Learning (AutoML)

\subsection{Structural Optimization of Bicycles Frames}
Structural considerations of a bicycle frame, such as geometry, material, and size can drastically affect the rider's experience. Typically, designers attempt to minimize the weight and cost of the frame, but removing too much material could increase the likelihood of structural failure, decrease the power transfer of pedaling into forward acceleration, or amplify the nerve-damaging effects of vibrations in intense bicycle riding. 

Since the inception of bicycles in the 1800s, designers have been steadily improving upon existing bicycles in search of increasingly optimal designs. Recent studies~\cite{de1999quantification, vanwalleghem2014development} have attempted to guide this incremental improvement through in-depth analysis of what physical features of a bicycle are most influential to the rider experience. In general, bicycles should be lightweight to allow fast acceleration and maneuverability, strong enough to resist failure under heavy loading, and stiff enough to maximize the power transfer of pedaling into acceleration. These conflicting objectives are not always intuitive, nor are they easily maximized.

Simulation of bicycle frames provides useful insight into how the rider experience can be improved. In one of the earliest attempts to numerically simulate bicycle frame loading using Finite Element Analysis (FEA), Soden~\etal~\cite{soden1986loads} represent a bicycle frame as a set of linear beams connected at a series of nodes to predict the deflections a frame might see under different riding conditions. Since then, with the development of advanced Computer-Aided Design (CAD) software and exponential growth in computational power, researchers have been able to represent more complex geometries~\cite{covill2015parametric}, develop more accurate estimates for stresses in bicycle frames~\cite{covill2016assessment}, and perform in-depth analyses of bicycle frame material selection~\cite{lessard1995utilization}.

Several studies have expanded on bicycle frame simulation with data-driven approaches to structural optimization of bicycles. For example, Cung \& Lee~\cite{chung2012parameters} simulate nearly 400 combinations of dimensional parameters for the four main tubes in the bicycle frame. They use these simulations to fit a model that determines the significance of each parameter in changing the structure of the bicycle frame. Cheng~\etal~\cite{cheng2016multi} use dynamic FEA simulations to optimize the bicycle frames for weight and impact resistance, with a focus on tube profiles. The authors simulate 18 designs chosen through a Design of Experiments and fit a surrogate model using Kriging. Other studies seek to optimize the bicycle frame by changing bicycle geometry.  Lin~\etal~\cite{lin2017structural} create a model that minimizes the deflection that a frame experiences under various loading conditions by changing the angles at which various tubes intersect. Covill~\etal~\cite{covill2014parametric} fit a regression model to capture how influential parameters affect bicycle frame deflection after simulating loading cases on 82 frames. 

Existing work at the intersection of numerical simulation and data-driven design for structural optimization of bicycles has shown great potential and paved the way for data-driven design to improve the rider experience. However, the existing body of work has a few gaps: The designs considered are typically selected through some randomized process and are sometimes unrealistic, their accuracy and generalizability are limited by the number of bikes simulated, and they are typically constrained to a small set of design parameters and load cases. Furthermore, studies seldom provide concrete design tools such as surrogate models, instead offering simple heuristics as key takeaways from their analysis. 

FRAMED drastically expands on this existing research. We simulate over 4500 bicycle frame models, considerably more than previous data-driven design studies, and release our dataset publicly for other researchers to use. These bicycle frame models are slightly modified from a collection of real bicycle designs, most of which are created by frame builders and enthusiasts. Models are parameterized by 37 design variables, constituting a significantly larger and more complex design space than used in previous studies. FRAMED also applies state-of-the-art Automated Machine Learning to fit high-performing surrogates on the dataset, which bicycle designers can use to estimate the performance of new frames. We introduce AutoML in the following section.

\subsection{Machine Learning and AutoML}

Datasets provide unique opportunities to develop Machine Learning-based surrogate models, which are frequently ``trained'' to learn the mapping from designs to performance values using example design-performance pairs. In this section, we introduce Machine Learning and Automated Machine Learning. Finally, we conclude the section with a brief overview of engineering design datasets. 

\subsubsection{Machine Learning (ML):} We provide a brief overview of Machine Learning, describing key terminology and methods, but refer readers to sources like~\cite{hastie2009elements} for a detailed and structured introduction. Though definitions vary, \textit{Machine Learning} is typically described as a subfield of \textit{Artificial Intelligence} spanning algorithms that improve automatically without explicit programming through the use of data. Predicting output variables based on input variables is a classic Machine Learning problem known as \textit{supervised learning}. Supervised learning is typically divided into \textit{regression}, which is the prediction of \textit{continuous} output variables, and \textit{classification}, which is the prediction of \textit{categorical} or \textit{discrete} variables. In supervised learning, an algorithm (model) gradually improves its predictive capability by studying example input-output pairs and adjusting its predictive mechanism when it makes mistakes, a process known as \textit{training}. 

We typically desire models to be \textit{generalizable}, i.e. when we query our model to predict outputs for inputs that it hasn't previously seen, our model should maintain its predictive performance. Generalizability is typically lost when a model \textit{overfits}, which can occur if the model only learns to accurately predict outputs for the exact training datapoints or similar data. To ensure generalizability, many data scientists use a process called \textit{cross-validation}. In cross-validation, models are evaluated on subsets of the dataset that are withheld from the model during training~\cite{hastie2009elements}. Evaluating the predictive performance of models on validation sets (data unused during training) is typically a much better reflection of a model's predictive performance on unseen data. 

\subsubsection{Automated Machine Learning (AutoML):} Selecting an optimal supervised learning algorithm and optimal training parameters (hyperparameters) for the selected model is highly dataset-dependent and constantly evolves as new methods are introduced. Model selection is often done through a combination of intuition and trial-and-error, however, this process is tedious and lacks rigor. A common approach to add rigor to the hyperparameter selection process involves performing optimization over the space of hyperperameters~\cite{feurer2019hyperparameter}. Since training is often tedious, optimization approaches like Bayesian Optimization that require relatively few test samples are typical choices~\cite{wu2019hyperparameter}. Recently, a procedure known as Automated Machine Learning (AutoML) has come to prominence. AutoML automates not only the selection of hyperparameters but also the selection of models and model architecture~\cite{he2021automl}. Many AutoML frameworks begin by processing the data and may perform automated feature engineering. After suitably processing the data, AutoML frameworks move on to algorithm selection, typically testing classic methods like Support Vector Machines and K-Nearest-Neighbors, as well as Neural Networks. After selecting several viable candidates, AutoML frameworks may go on to optimize one or several of the algorithms selected. Typically, this involves optimization of hyperparameters, but in the case of Neural Networks, this may also involve Neural Architecture Search~\cite{ren2021comprehensive}. AutoML lowers the bar of entry for Machine Learning, making it accessible to those without intuition and training, and saving practitioners time and money. Furthermore, AutoML frameworks routinely outperform experienced data scientists in identifying optimal supervised learning models~\cite{hutter2019automated}. 

\subsubsection{AutoML in Engineering Design}
Machine Learning has been growing in prominence within engineering and countless researchers have applied ML to design-related problems. Surrogate modeling is one of the most prevalent applications of Machine Learning in Engineering Design. For a detailed review of surrogate modeling in design using non-AutoML approaches, we refer readers to reviews like ~\cite{alizadeh2020managing, sun2019review, viana2021surrogate}. While applications in engineering design have typically trailed methodological advancements in Machine Learning by several years, Automated Machine Learning has thus far seen limited use in engineering design, despite its established dominance in other fields. Instead of leveraging AutoML to identify high-performing algorithms systematically, design practitioners often simply apply their favorite Machine Learning technique to their problems. Active engineering design research disciplines such as materials design~\cite{wei2019machine, guo2021artificial}, biosystems design~\cite{volk2020biosystems}, structural design~\cite{sun2021machine, thai2022machine}, and additive manufacturing~\cite{goh2021review, razvi2019review} have seen minimal use of AutoML in practice. Given its ability to identify optimal ML models with limited user expertise and intuition, AutoML is a technique that many design practitioners with limited data science experience can leverage to great effect. 

\subsubsection {AutoGluon:} While many AutoML methods are limited to finding a single optimal class of algorithm as well as the associated hyperparameters to maximize performance, several AutoML frameworks have proposed approaches to identify even better-performing models. One such approach, AutoGluon, proposes a novel layer-stack ensembling approach, leveraging the known tendency of ensemble predictors to outperform individual models~\cite{erickson2020autogluon}. AutoGluon also utilizes k-fold bagging, an extension of cross-validation which allows all of the available data to be used during training with minimal risk of overfitting. For more details about AutoGluon, we refer the reader to~\cite{erickson2020autogluon}. 
In this work, we apply AutoGluon to the bicycle frame structural performance and feasibility prediction problems. Due to AutoML's limited use and relatively unproven performance in engineering design applications, we validate AutoGluon's performance against models selected through Bayesian hyperparameter optimization. 

\subsection{Datasets in Engineering Design}
Advancements in applied Machine Learning are enabled by the datasets and problems to which they are applied. Though data-driven automation is extremely desirable in design, high-quality design datasets are particularly difficult to come by. While computer vision datasets with millions of images are relatively common, design datasets with even thousands of entries are few and far between. We briefly mention several noteworthy engineering design datasets here but refer readers to the more detailed review in ~\cite{regenwetter2021deep}. In particular, we focus on design datasets with associated engineering performance data. The UIUC Airfoil Database is a dataset of nearly 1600 real airfoil designs and is extended by Chen~\etal~\cite{chen2019aerodynamic} to include aerodynamic lift and drag performance values. Wang~\etal~\cite{wang2020deep} introduce a microstructure dataset with associated tensor stiffness values. Wollstadt~\etal~\cite{wollstadt2022carhoods10k} introduce a dataset of 10,000 car hood topologies with associated structural performance data. In addition to the aforementioned datasets, numerous microstructure, topology optimization, and molecular design datasets often include associated material or chemical property values and are frequently used in structural, material, and biochemical design problems~\cite{nie2021topologygan, zhao2018nanomine}. 

\section{Methodology}
In this section, we discuss the various methodology decisions behind the dataset including design parameterization, modeling, analysis of geometric feasibility, load cases, material selection, and meshing. 

\subsection{Parameterization and Modeling}
\begin{figure}[!htb]
    \centering
    \includegraphics[width=0.9\textwidth]{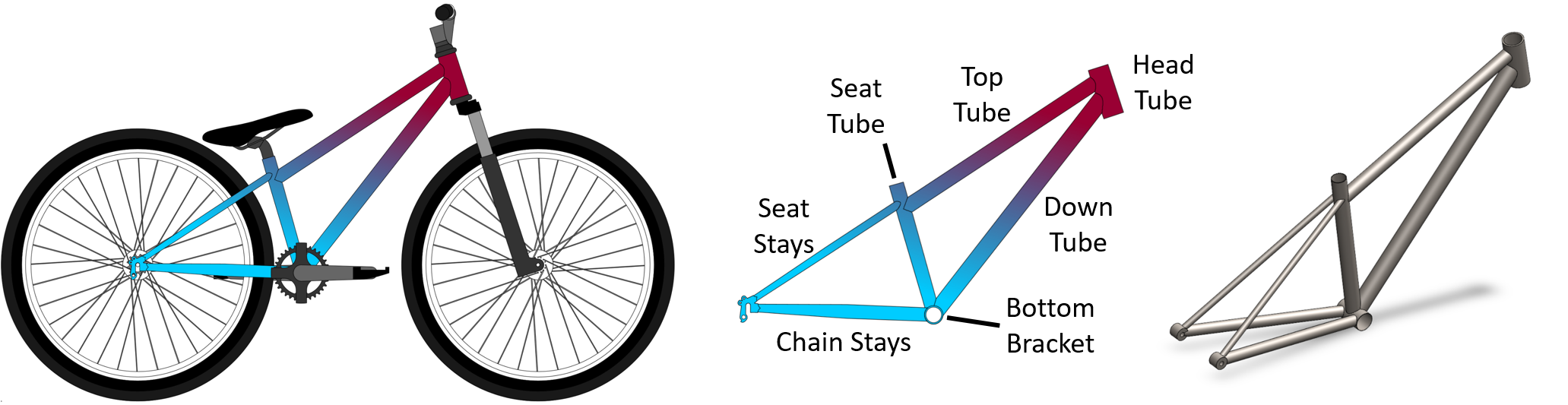}
    \caption{Side-by-side comparison of bicycle from BIKED, annotated frame, and 3D model from FRAMED.}
    \label{fig:Frame2Model}
\end{figure}

In generating our dataset, we utilize models from BIKED~\cite{regenwetter2022biked}, a dataset comprised of 4500 individually designed bicycle models sourced from hundreds of designers who use the BIKECAD software.
The BIKED dataset contains over 1300 design parameters, roughly 200 of which we identify as being directly related to the bicycle frame. To reduce the design space and ensure that 3D models can be reliably built from these design parameters, we make several key simplifications to these bicycle frame models:
\begin{enumerate}
    \item We only consider the ``diamond'' bicycle frame topology, such as the frame shown in Figure~\ref{fig:Frame2Model}. 
    \item We assume all tubes have a uniform cross-section along their length and are straight
    \item We do not consider rounded junctions or fillets at the intersections of tubes
\end{enumerate}

These simplifications allow us to reduce the design space to 37 parameters from the original 200 frame-related parameters in BIKED. Most of these parameters are taken directly from BIKED, while a few are calculated deterministically by combining multiple BIKED design parameters. These 37 parameters can roughly be broken down into several groups, such as tube diameters, tube thicknesses, and dimensions of the high-level frame geometry. Additionally, we maintain two parameters from BIKED which serve as boolean flags indicating whether or not the frame has chain stay or seat stay bridges (bridges are crosspieces between the stays that add support). Finally, we use a single material parameter, which is discussed in more detail below in Section~\ref{materials}. A summary of the parameter types is included in Table~\ref{tab:params}. A side-by-side comparison of an original BIKED bicycle model, the same BIKED model with the frame isolated, and the corresponding 3D model generated based on this BIKED model is shown in Figure~\ref{fig:Frame2Model}.

\begin{table}[!htb]
\caption{Summary of parameters used to represent the bicycle frame design space.}
\label{tab:params}
\centering
\begin{tabular}{|l|c|c|l}
\cline{1-3}
\textbf{Parameter Type} & \textbf{Data Type} & \textbf{Count} &  \\ \cline{1-3}
Frame Geometry Relations & Continuous & 18 &  \\ \cline{1-3}
Tube Outer Diameters & Continuous & 9 &  \\ \cline{1-3}
Tube Thicknesses & Continuous & 7 &  \\ \cline{1-3}
Frame Material & Categorical & 1 &  \\ \cline{1-3}
Seat/Chain Stay Bridge Flags & Boolean & 2 &  \\ \cline{1-3}
\textbf{Total} & \multicolumn{1}{l|}{} & \textbf{37} &  \\ \cline{1-3}
\end{tabular}
\end{table}

\begin{figure}[!htb]
    \centering
    \includegraphics[width=0.9\textwidth]{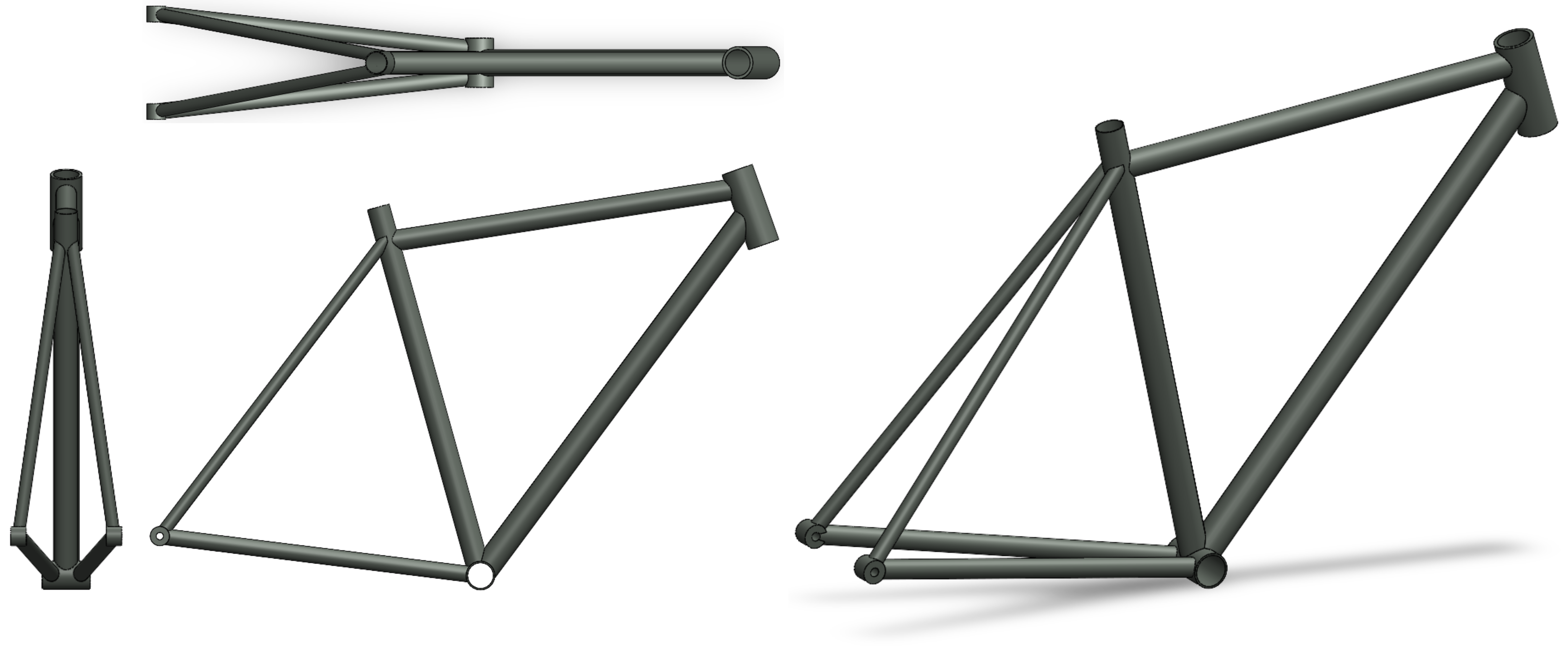}
    \caption{Different views of example bicycle frame model.}
    \label{fig:model}
\end{figure}

One of the key limitations acknowledged by BIKED's authors is the limited diversity present in certain design parameters, largely due to peculiarities stemming from the BikeCAD software from which designs were sourced. BikeCAD has no 3D modeling feature or inbuilt simulation capabilities, so many parameters adding ``depth'' to the model are largely irrelevant in BikeCAD designs. This issue is especially pronounced in the tube thicknesses, with over 99\% of all models having the same tube thickness values. To promote diversity in our dataset, we manually override the seven tube thickness parameters with randomly sampled thicknesses. We sample a 7D vector from a Sobol sequence, then logarithmically scale these vectors in an element-wise fashion to a range of 0.5-10 mm. The resulting bike models' seven tube thickness values randomly lie between 0.5 and 10 mm, with a bias towards thinner tubes. 

To automate the generation of 3D models to simulate, we create an adaptive 3D template model. This model uses a set of predefined equations to deterministically calculate key dimensions based on our 37-parameter design representation. The model then leverages the inbuilt constraint-resolution functionality of SolidWorks to build a final 3D model for each frame. Figure~\ref{fig:model} shows several views of a road bike frame model build with this method. 
\subsection{Geometric Feasibility}

Our 37-variable parameterization makes for a diverse design space but also introduces possibilities for infeasible combinations of parameters. To avoid geometrically infeasible models, we implement a list of geometric ``checks.'' A few of these checks are listed below:
\begin{itemize}
    \item Tube thicknesses, diameters, and lengths must be positive
    \item Seat stays and chain stays must intersect with the seat tube and bottom bracket respectively.
    \item Head tube and seat tube angles are between 0 and 180 degrees. 
\end{itemize}

We find that 222 of BIKED's 4512 models fail these explicit feasibility checks. Despite these checks, 172 bicycle models still fail to build correctly when the parameters are fed into the adaptive 3D frame model, possibly due to geometric infeasibilities. Hence, 4118 models are found to be geometrically valid. 

\subsection{Load Cases and Simulation Setup}

\begin{figure*}[ht]
    \centering
    \includegraphics[width=\textwidth]{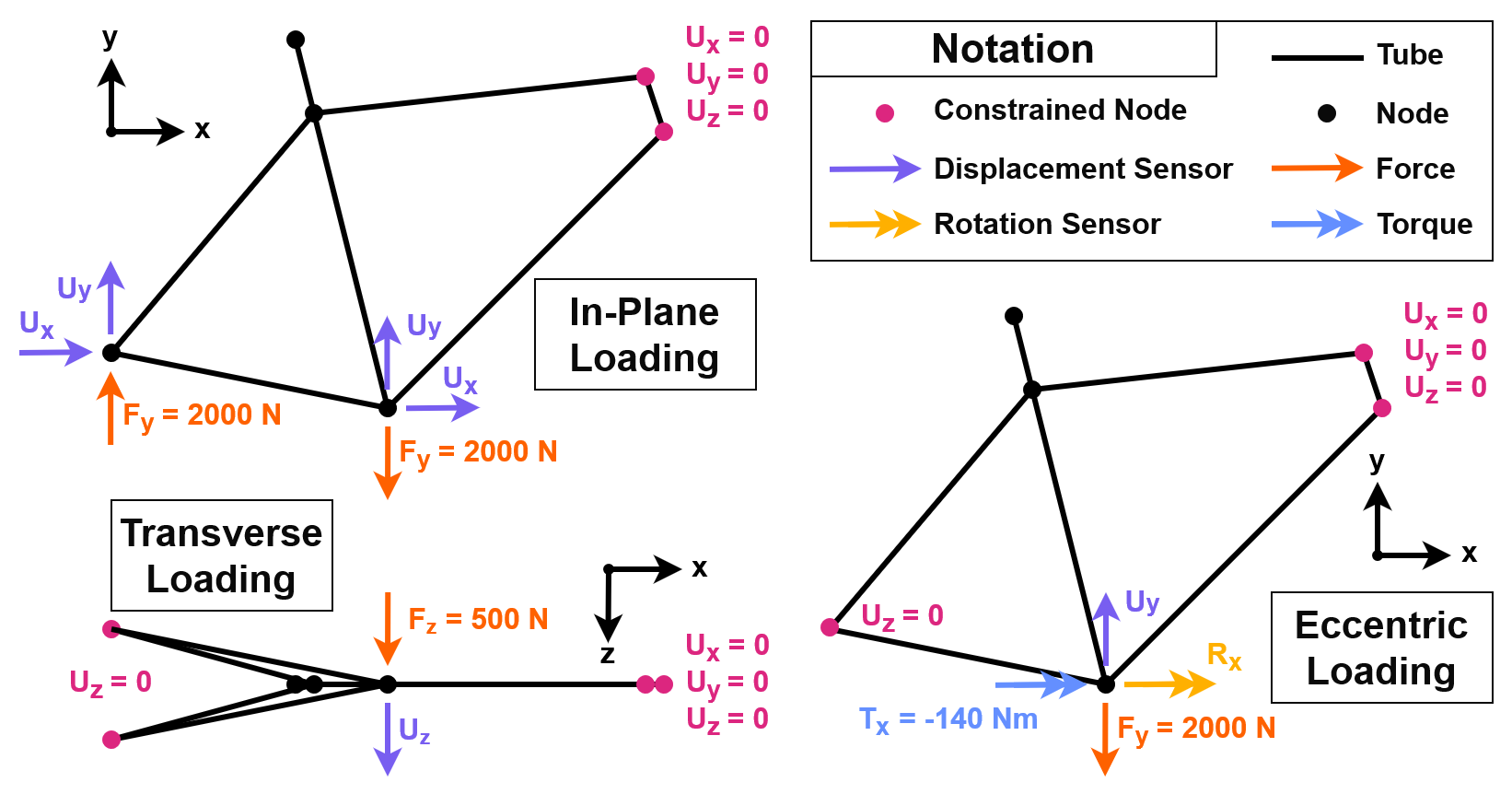}
    \caption{Diagrams of the three simulations designed to test frames during in-plane, transverse, and eccentric load cases.}
    \label{fig:loading}
\end{figure*}

We seek to develop a concise set of tests that effectively evaluate a wide variety of structural considerations of the bicycle frame. We follow the methodology proposed in Vanwelleghem~\etal~\cite{vanwalleghem2014development} to evaluate in-plane, transverse, and eccentric stiffness. The authors propose three load cases to evaluate bicycle frames. Though Vanwelleghem~\etal~focus only on stiffness and don't specify load magnitudes in their methodology, we require loads to roughly estimate maximum stresses and safety factors. Soden~\etal~\cite{soden1986loads} study forces applied to the bicycle during actual ridership in several road racing conditions (starting, climbing, braking,~\etc) and find a maximum pedal force of $1447~N$ across these conditions. We select loading magnitudes based on these findings and illustrate our load cases in Figure~\ref{fig:loading}. Based on these studies and domain knowledge, we introduce three load cases that are applied to every bicycle frame. These cases are defined as follows:
\begin{enumerate}
    \item In-Plane Stiffness: We apply $2000~N$ upwards to the dropouts and $2000~N$ downwards to the bottom bracket while holding the head tube fixed. We measure vertical and lateral displacements at the bottom bracket and dropouts as well as the safety factor. This load case corresponds to a `normal riding' scenario, where large in-plane forces may be caused by uneven riding surfaces.  
    \item Transverse Stiffness: We apply $500~N$ laterally to the bottom bracket while holding the head tube fixed and preventing lateral deflection at the dropouts. We measure lateral displacement at the bottom bracket. Though transverse loading is relatively small during normal cycling, deflection can still be significant and contribute to reduced power efficiency.
    \item Eccentric Stiffness: We apply a $2000~N$ downwards force and $140~Nm$ moment to the bottom bracket (representing a $2000~N$ force applied at an offset of $7~cm$ from the bottom bracket). We measure vertical and angular displacement of the bottom bracket as well as the safety factor. This load case roughly captures an acceleration-from-stationary scenario where the rider applies their entire body weight (or more if they are pulling up on the handlebars) to a single pedal. 
\end{enumerate}

Six displacements and one rotation are measured across the three loading cases, which can be used to find various directional and rotational stiffnesses of the frame. Additionally, safety factors are measured for Simulations 1 and 3. Finally, we also measure the weight of the frame model. We note that these forces may not cover extreme use cases. For example, De~Lorenzo~\etal~\cite{de1999quantification} study forces on a bicycle during ``aggressive off-road cycling'', including a $2.5~m$ jump and find a maximum rear wheel loading of $4000~N$.

\subsection{Material Properties} \label{materials}
\begin{figure}[!htb]
    \centering
    \includegraphics[width=0.8\textwidth]{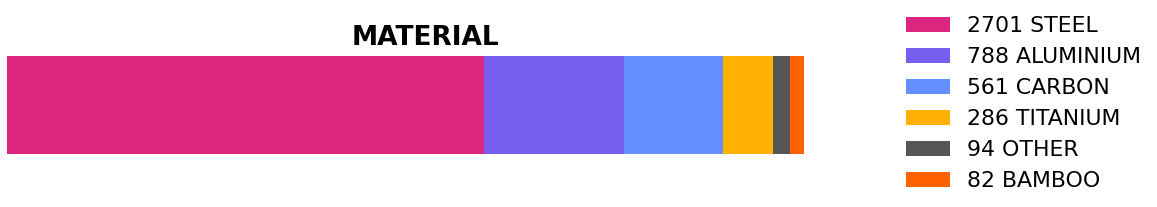}
    \caption{Breakdown of bikes by material in original BIKED data.}
    \label{fig:materials}
\end{figure}

BIKED provides a categorical ``material'' parameter consisting of one of six material classes. The breakdown of bicycle frames by material in the original dataset is shown in Figure~\ref{fig:materials}. Three of these (steel, aluminum, and titanium) are isotropic\footnote{We acknowledge that manufacturing processes, including those used for bicycle frame tubing, can introduce anisotropic properties in a material. Modeling anisotropic behavior was deemed to be outside of the scope of this work, and we acknowledge that this simplification may cause minor inaccuracies.} while carbon and bamboo are anisotropic. Since anisotropic materials are difficult to simulate without additional information about material orientation, we replace bamboo and carbon fiber, as well as the unspecified ``other'' category with aluminum. BIKED does not specify the alloy of steel, aluminum, and titanium used in bicycle models. Therefore, we select material properties of steel, aluminum, and titanium that are generally characteristic of common bicycle tube alloys. We select steel properties common for a heat-treated chrome-molybdenum steel such as AISI 4130 Steel, which is fairly representative of the steels used in bicycle fabrication. We select aluminum and titanium properties of 6061-T6 aluminum and Ti-6Al-4V, respectively, which are two of the most commonly used alloys in the industry. These properties are summarized in Table~\ref{tab:materials}. 

\begin{table}[!htb]
.\caption{Selected material properties for steel, aluminum, and titanium used in simulation}
\label{tab:materials}
\centering
\begin{tabular}{lccc}
\hline
\multicolumn{1}{|l|}{Material}                          & \multicolumn{1}{c|}{\textbf{Steel}} & \multicolumn{1}{c|}{\textbf{Aluminum}} & \multicolumn{1}{c|}{\textbf{Titanium}} \\ \hline
\multicolumn{1}{|l|}{Elastic Modulus ($GPa$)}             & \multicolumn{1}{c|}{205}            & \multicolumn{1}{c|}{69}                & \multicolumn{1}{c|}{105}               \\ \hline
\multicolumn{1}{|l|}{Poisson's Ratio}                   & \multicolumn{1}{c|}{0.285}          & \multicolumn{1}{c|}{0.330}             & \multicolumn{1}{c|}{0.310}             \\ \hline
\multicolumn{1}{|l|}{Shear Modulus ($GPa$)}               & \multicolumn{1}{c|}{80}             & \multicolumn{1}{c|}{26}                & \multicolumn{1}{c|}{41}                \\ \hline
\multicolumn{1}{|l|}{Density ($kg/m^3$)} & \multicolumn{1}{c|}{7850}           & \multicolumn{1}{c|}{2700}              & \multicolumn{1}{c|}{4429}              \\ \hline
\multicolumn{1}{|l|}{Tensile Strength ($MPa$)}            & \multicolumn{1}{c|}{731}            & \multicolumn{1}{c|}{310}               & \multicolumn{1}{c|}{1050}              \\ \hline
\multicolumn{1}{|l|}{Yield Strength ($MPa$)}              & \multicolumn{1}{c|}{460}            & \multicolumn{1}{c|}{275}               & \multicolumn{1}{c|}{827}               \\ \hline
& \multicolumn{1}{l}{}                & \multicolumn{1}{l}{}                   & \multicolumn{1}{l}{}                   \\
& \multicolumn{1}{l}{}                & \multicolumn{1}{l}{}                   & \multicolumn{1}{l}{}                   \\
& \multicolumn{1}{l}{}                & \multicolumn{1}{l}{}                   & \multicolumn{1}{l}{}                  
\end{tabular}

\end{table}

\subsection{Mesh Resolution}
\begin{figure*}[!htb]
    \centering
    \includegraphics[width=\textwidth]{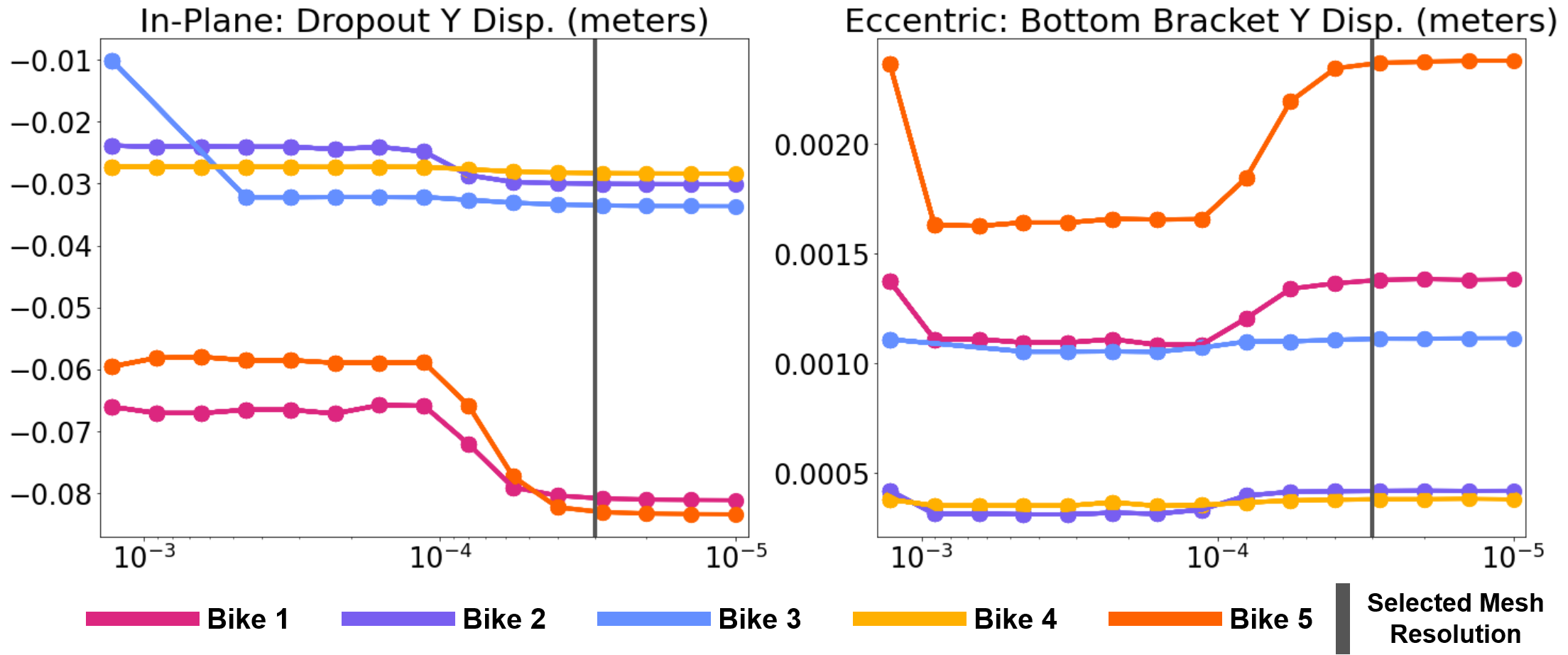}
    \caption{Convergence study calculating two quantities of interest at different mesh resolutions across five different bicycle frame models. Mesh resolution is shown on horizontal axes and is measured in meters. The selected mesh resolution of 0.03 mm is indicated on the plots.}
    \label{fig:convergence}
\end{figure*}

In numerical simulations, mesh resolution is an essential parameter that balances the tradeoff between computational cost and simulation fidelity. Since this work simulates thousands of models, appropriately balancing computational cost and fidelity is essential. To study this balance, we randomly select five bicycle frame models to test in each of our three simulation setups. For each study, we test a logarithmic sweep of mesh resolutions with minimum cell size ranging from $0.01~mm$ to $1.28~mm$. Meshes are generated using SolidWorks' ``Blended curvature-based mesh.'' In each test, the maximum cell size was set to 100 times the minimum cell size, and the cell growth ratio between adjacent cells was set to 1.3. We examined convergence across mesh resolutions for each of our ten parameters of interest and documented two sample plots in Figure~\ref{fig:convergence}. 

Although displacement values stabilize at fine mesh resolution, we observe in our studies that safety factors do not perfectly stabilize at even the finest of mesh resolutions tested. Qualitative analysis of simulation results indicates that the safety factors are reflecting extreme local stress concentrations at the junctions of the tubes. Thus, the low safety factors at finer resolutions can likely be attributed to the imperfect modeling of the bicycle frame, particularly the infinite curvature at the tube junctions in the model. As such, we caution users of the dataset to expect some error in reported stress and safety factor values. 

In general, displacement values are stable for mesh resolutions between $0.16~mm$ and $0.32~mm$. Above $0.64~mm$, displacement values are relatively unstable and simulations occasionally fail to converge. Displacement values are also relatively unstable for mesh resolutions between $0.04~mm$ and $0.16~mm$. We hypothesize that this range of mesh resolutions critically impacts fidelity since tube thicknesses may be as small as $0.5~mm$ and an accurate simulation should place several cells spanning the thickness of any key geometry. Below $0.04~mm$, displacements are fairly stable.

We select a mesh resolution of $0.03 mm$ for our simulations to attain a reasonably precise estimate of displacements while avoiding the extreme cost brought about by finer meshing. 

\section{Validation}

To demonstrate that our frame model and meshing setup yield meaningful simulation results, we validate against existing published data. Validation using physical testing is usually a costly and time-consuming method but is often the most rigorous. Fortunately, many existing studies have published results of physical experimentation on bicycle frames. Few of these, however, publish enough details on parametric data about the bikes they test for us to construct an accurate 3D bicycle frame model for simulation. We select a 1996 study by Damon Rinard~\cite{rinard_1996} which physically tested over 70 bicycle frames for transverse deflection of the front and rear triangles. From their study, we select three frames for which we were able to find sufficient parametric data to approximate the 3D frame models: the DeRosa SLX, Casati Gold Line, and Holland SL/SP. Much of the parametric data comes from~\cite{equus_bicycle}, which also provides estimates for frame mass. We mimic Rinard's loading and measurement setup and compare simulated deflection values with reported values as well as frame model mass with reported frame mass. These results are presented in Table~\ref{tab:validation}.

\begin{table}[!htb]
\centering
\caption{Physical validation study demonstrates modest error in FEA simulation results.}
\label{tab:validation}
\resizebox{\textwidth}{!}{%
\begin{tabular}{|l|ccc|ccc|ccc|}
\hline
 & \multicolumn{3}{c|}{DeRosa SLX} & \multicolumn{3}{c|}{Casati Gold Line} & \multicolumn{3}{c|}{Holland SL/SP} \\ \hline
 & \multicolumn{1}{l|}{Front} & \multicolumn{1}{l|}{Rear} & \multicolumn{1}{l|}{Model} & \multicolumn{1}{l|}{Front} & \multicolumn{1}{l|}{Rear} & \multicolumn{1}{l|}{Model} & \multicolumn{1}{l|}{Front} & \multicolumn{1}{l|}{Rear} & \multicolumn{1}{l|}{Model} \\
 & \multicolumn{1}{l|}{Defl.} & \multicolumn{1}{l|}{Defl.} & \multicolumn{1}{l|}{Mass} & \multicolumn{1}{l|}{Defl.} & \multicolumn{1}{l|}{Defl.} & \multicolumn{1}{l|}{Mass} & \multicolumn{1}{l|}{Defl.} & \multicolumn{1}{l|}{Defl.} & \multicolumn{1}{l|}{Mass} \\ \hline
Actual & 0.40 & 0.15 & 1.966 & 0.44 & 0.15 & 1.966 & 0.38 & 0.13 & 1.962 \\
Simulated & 0.297 & 0.116 & 1.69 & 0.3028 & 0.124 & 1.80 & 0.26 & 0.107 & 1.77 \\ \hline
Error & 26\% & 23\% & 14\% & 31\% & 17\% & 8\% & 32\% & 18\% & 10\% \\ \hline
\end{tabular}%
}
\end{table}
The comparison shows that our simulations have similar trends of deflection and mass as Rinard's studies. However, some discrepancy is often expected between simulation and real-world testing and we find that our simulations tend to underestimate the front and rear deflections compared to Rinard's reported values. There are several potential explanations for the difference, a few of which we discuss here.
First, the discrepancy in mass can largely be explained by the fact that our model does not include the frame's fork while the experimental values do, explaining the underestimation of mass values. 
Second, we suspect that measured deflections in Rinard's studies fail to eliminate deflection caused by the compliance of their clamping scheme. This likely explains why the simulated values for front and rear deflection consistently show similar errors across all three bikes tested. Since the moment on the clamping mechanism is roughly the same for each test (the distances from the clamp to the front and rear ends of the bicycle do not vary much from bike to bike), compliance in the clamp would contribute roughly the same error. For these reasons, we are optimistic that our simulations are even closer to real-world behavior than this validation study would indicate. In future work, we plan to conduct our own physical validation to more accurately validate our simulation results. Nonetheless, given certain simplifications made in modeling (ignoring fillets, approximating material properties with the values for typical alloys, etc.), we consider these error values to be acceptable. 


\section{Dataset Analysis}

\subsection{Exploring the Performance Space}
\begin{figure*}[!htb]
    \centering
    \includegraphics[width=.99\linewidth]{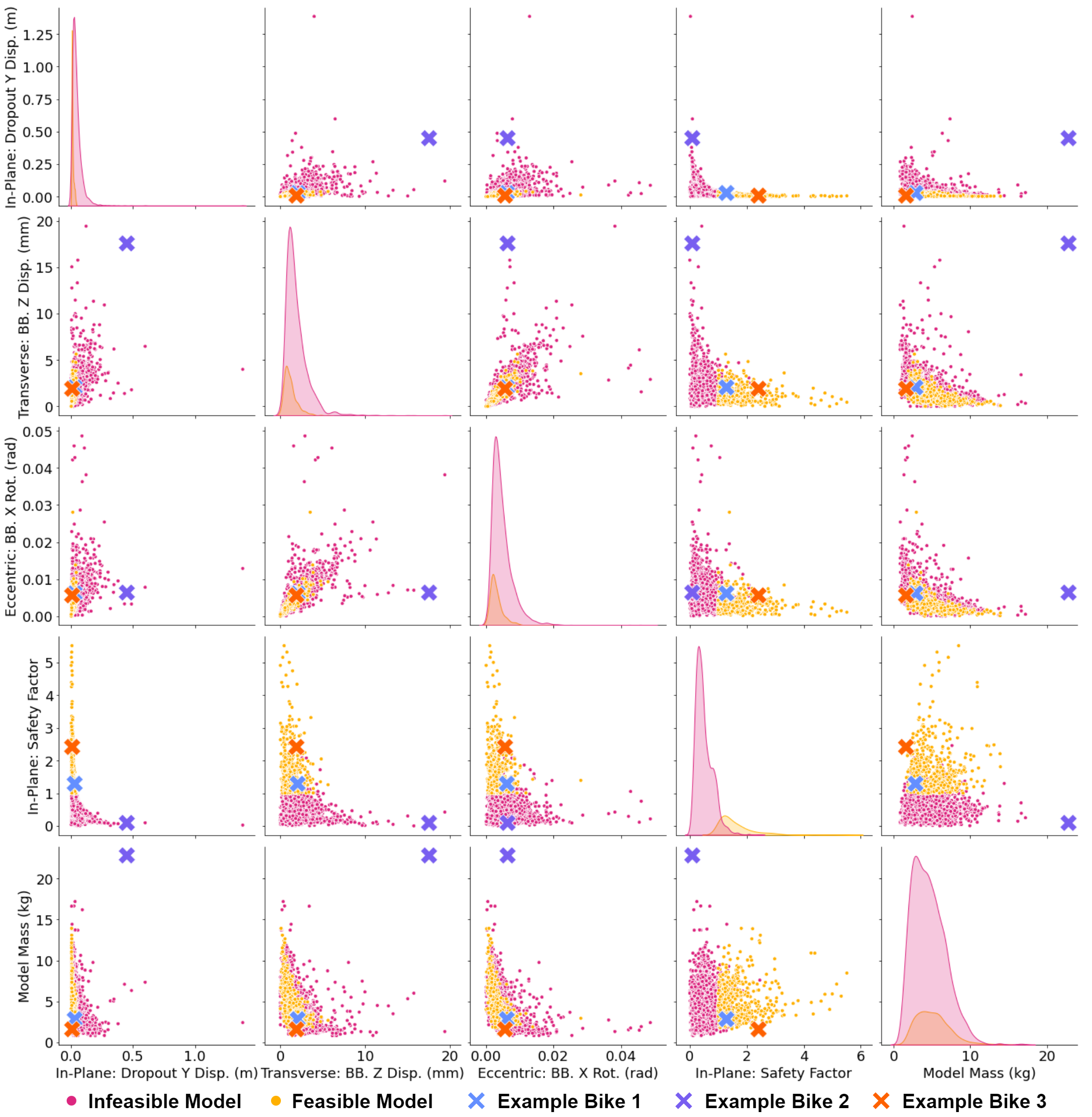}
    \caption{Plot showing: 1) Kernel Density Estimate plots (diagonal plots). 2) Scatterplots over each pair of performance parameters (off-diagonal plots). 3) Classification of bicycle frame models into feasible and infeasible models. 4) Three example frames that we discuss as case studies.}
    \label{fig:Validity_Pairplot}
\end{figure*}

Through our simulations, we captured ten structural performance values for each of the 4118 geometrically valid bicycle frames. Using a minimum Factor of Safety (FoS) threshold of 1.0, we can immediately identify bicycles that fail under reasonable load cases. 3419 of 4118 frames simulated fail under at least one of the loading cases, a reminder of the difficult balance of parameters and the complexity of the bicycle design problem. Designers may often not anticipate that a particular bicycle design is structurally deficient until physically testing the frame. To make for easier visualization, we explore the design space with a subset of five of the ten performance values: Dropout displacement during in-plane loading, bottom bracket displacement during transverse loading, bottom bracket rotation during eccentric loading, safety factor during in-plane loading, and weight. Figure~\ref{fig:Validity_Pairplot} shows a visualization of this subset, with Kernel Density Estimate plots over each performance parameter and scatterplots over each pair of performance parameters. Note that we take absolute values of deflection values (Simulation 1 deflections can be positive or negative). Additionally, points and histograms are organized based on bicycle frame model validity. In this case, we take a frame model to be valid if both safety factor values measured (one not shown) are greater than 1. Additionally, we label three bicycle frames on these plots to analyze in the following section. Based on these plots, we can make several observations. For example, looking at these histograms, we see that valid bicycle frames tend to have much smaller deflections than invalid frames. We can also see that the two distributions over mass align very closely. Based on the scatterplots, we can also observe some correlations between objectives. For example, heavier models tend to have deflections with smaller magnitudes.

\subsection{Case Studies}
\begin{figure}
     \centering
     \begin{subfigure}[b]{0.3\textwidth}
         \centering
         \includegraphics[width=\textwidth]{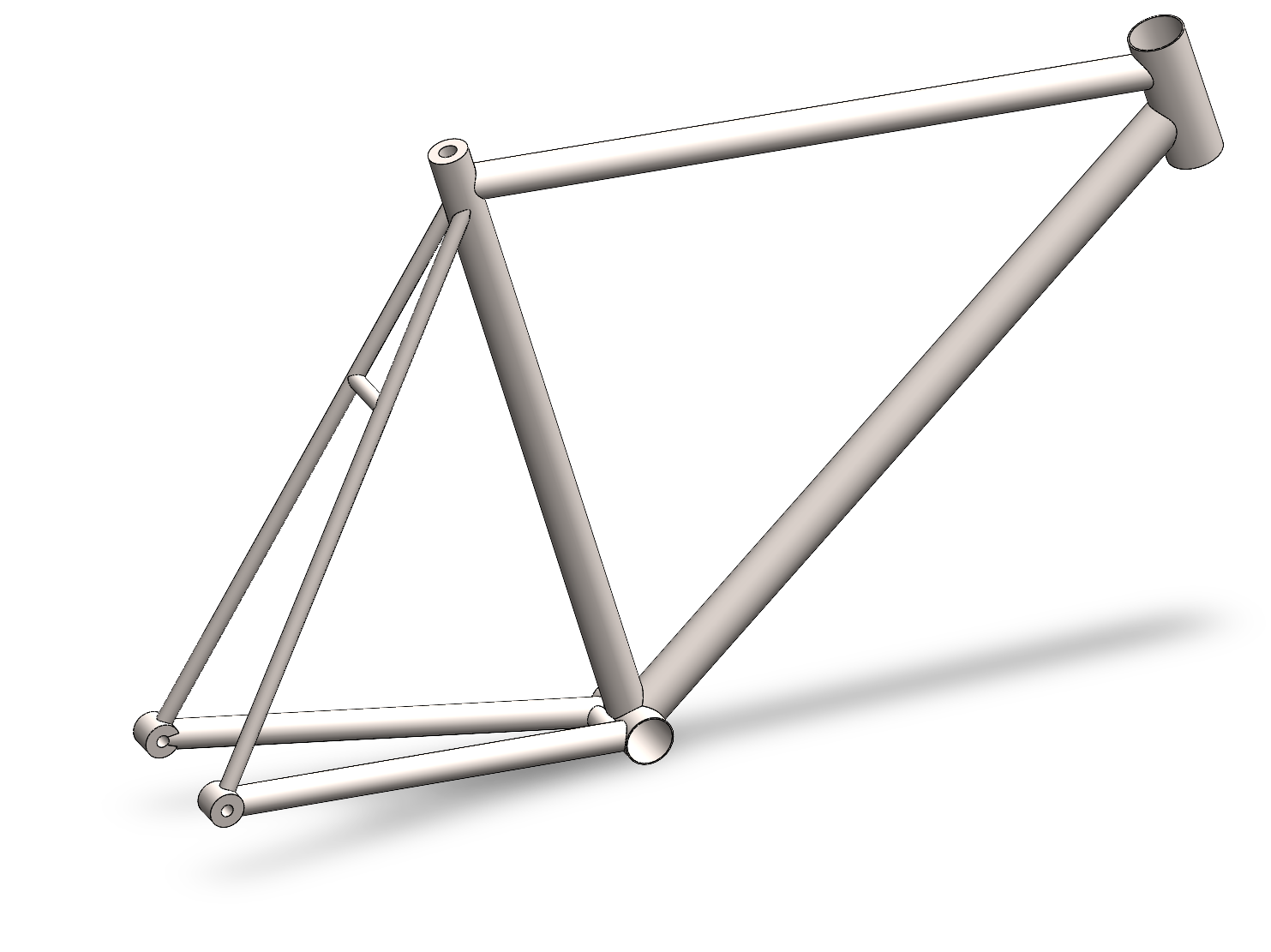}
         \caption{Example frame 1 (Valid)}
         \label{fig:EB1}
     \end{subfigure}
     \hfill
     \begin{subfigure}[b]{0.3\textwidth}
         \centering
         \includegraphics[width=\textwidth]{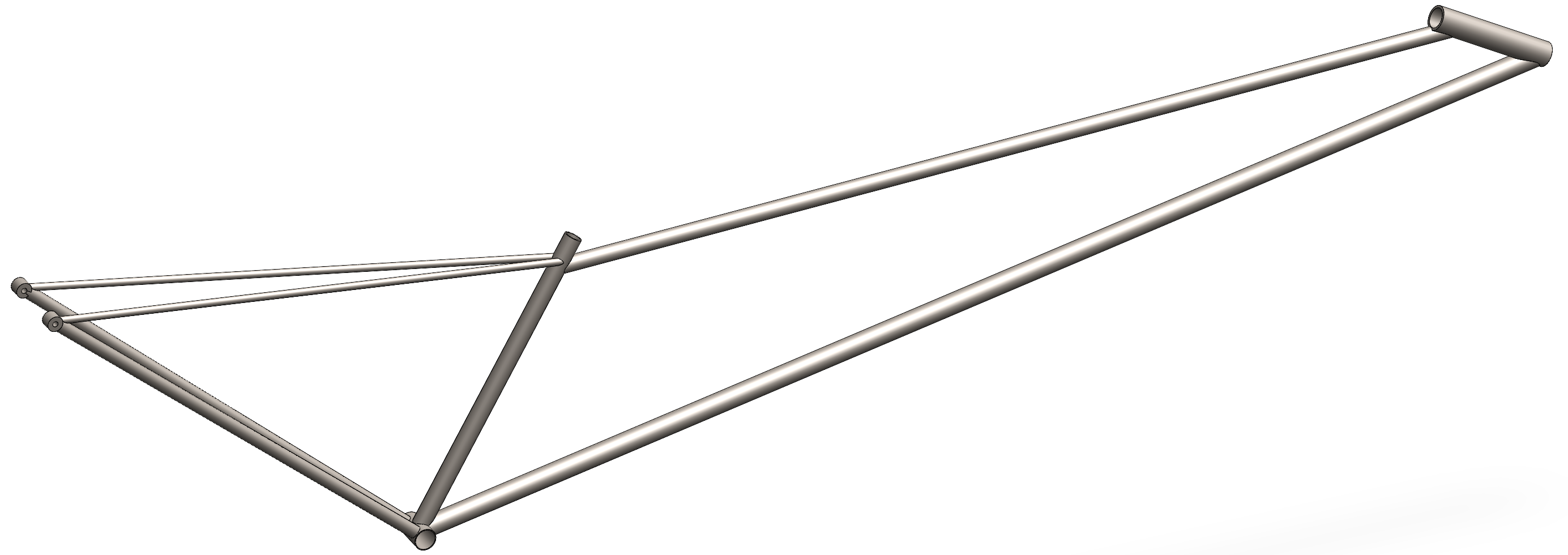}
         \caption{Example frame 2 (Invalid)}
         \label{fig:EB3}
     \end{subfigure}
     \hfill
     \begin{subfigure}[b]{0.3\textwidth}
         \centering
         \includegraphics[width=\textwidth]{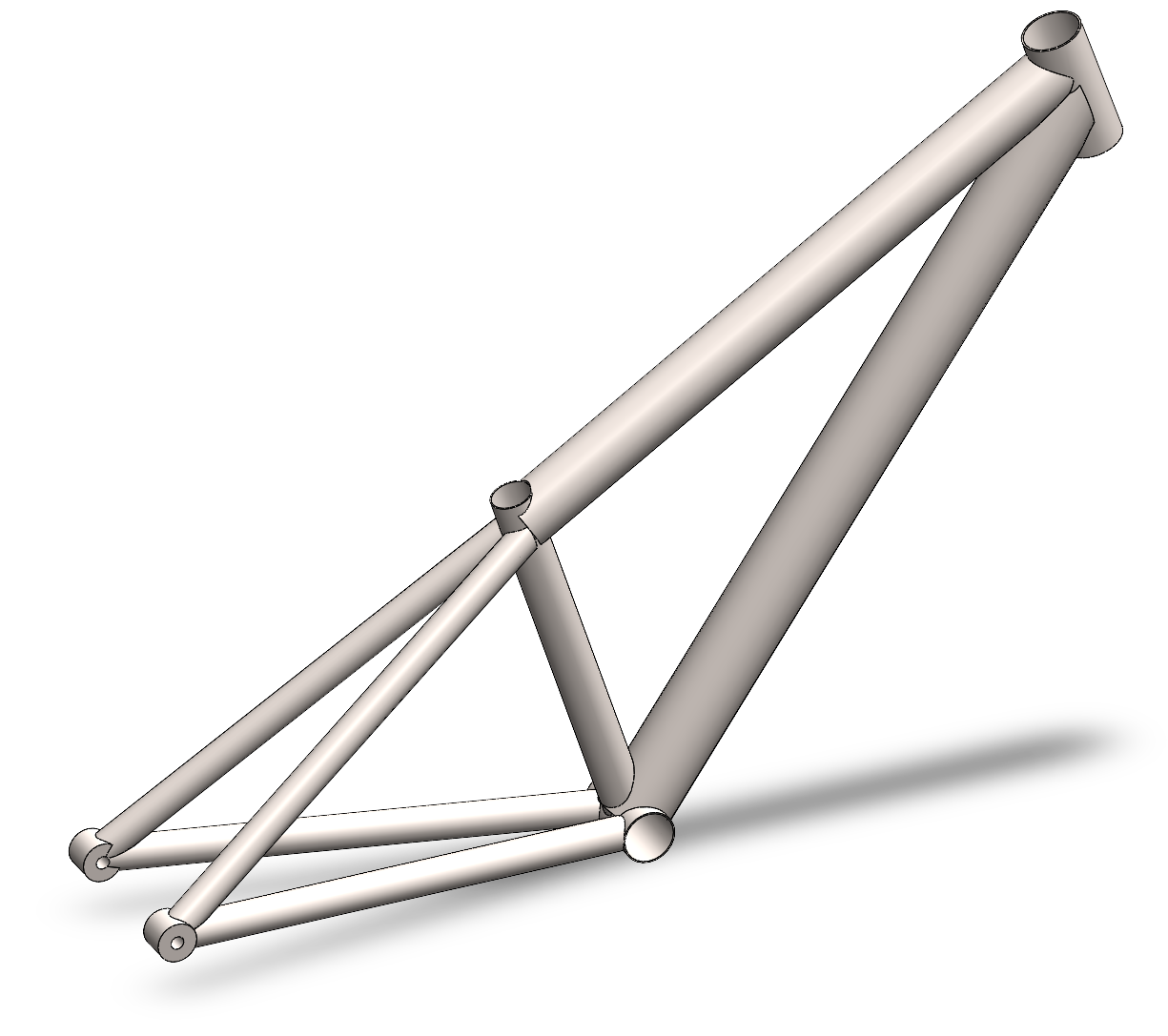}
         \caption{Example frame 3 (Valid)}
         \label{fig:EB2}
     \end{subfigure}

        \caption{Highlighted example frames, showing the frame with the highest vertical deflection (left), the frame with the highest safety factor (middle), and the lightest valid frame (right).}
        \label{fig:EBs}
\end{figure}

Here, we examine three sample frames to provide potential users with some intuition about the dataset. We demonstrate a `typical' bike, but also examine some outliers in the data, such as artistic bikes or bikes designed for children. Three frames are shown in Figure~\ref{fig:EBs} (corresponding to the bikes indicated in Figure~\ref{fig:Validity_Pairplot}) and are discussed below. We recommend that users be aware of the broad variety of use cases, bike styles, and user groups represented in the data and refer to~\cite{regenwetter2022biked} for more details. 

\subsubsection{Frame 1:} This road bike frame was randomly selected from the valid bikes. Though it contains some eccentricities such as an unconventionally thick seat tube and chain stay bridge fused with the bottom bracket, it exemplifies a relatively `well-designed,' albeit fairly typical frame. Weighing in at $2.84~kg$, its safety factors are 1.3 and 1.1 in Simulations 1 and 3 respectively. The frame's deflections are rather moderate across all load cases. 
\subsubsection{Frame 2:} This frame was manually selected as the heaviest frame in the dataset, weighing $22.83~kg$. This highly unconventional frame is nearly three meters long and, judging by the original design in BIKED, appears to have been designed as an `artistic' piece, rather than a functional bike. It was designed to support massive wheels with a diameter of $1.75~m$ extending far above the rider's head. Despite its heavy steel frame, this bike undergoes extreme stress and deflections during simulation and has a safety factor of less than 0.1. Unconventional frames like this raise the diversity of the dataset and often stand out as outliers in the performance space, but require users to be cognizant of the varied use cases represented in the data. 

\subsubsection{Frame 3:} This frame was identified as the lightest valid frame in the dataset at $1.6~kg$. Not only is the frame structurally valid, but it also boasts safety factors of 2.42 and 2.99 in Simulations 1 and 3 and some of the smallest deflections in the dataset. While the frame utilizes lightweight titanium and features thin wide tubes that effectively use their higher second moment of area to minimize bending stress and deflection, its most notable feature is its size. This bike is extremely compact and may have been designed for children. We caution users interested in performing direct optimization of structural performance: Appropriate constraints are essential, as `optimality' can be trivially attained by letting the dimensions of the frame approach zero, as this example illustrates.

\section{Predicting Structural Performance using Surrogate Models}
Given a new bicycle design concept, a designer might like to quickly gauge its structural performance, without building a costly and time-consuming prototype. Instead of creating a digital model and running numerical simulations, which are typically tedious, designers can leverage surrogate models to estimate performance in fractions of a second. In this section, we demonstrate that FRAMED can be used to train algorithms to rapidly and accurately predict performance and feasibility for previously unseen bicycle frame designs. 

\subsection{Selecting a Model using Automated Machine Learning (AutoML)}
\begin{figure*}[!htb]
    \centering
    \includegraphics[width=\textwidth]{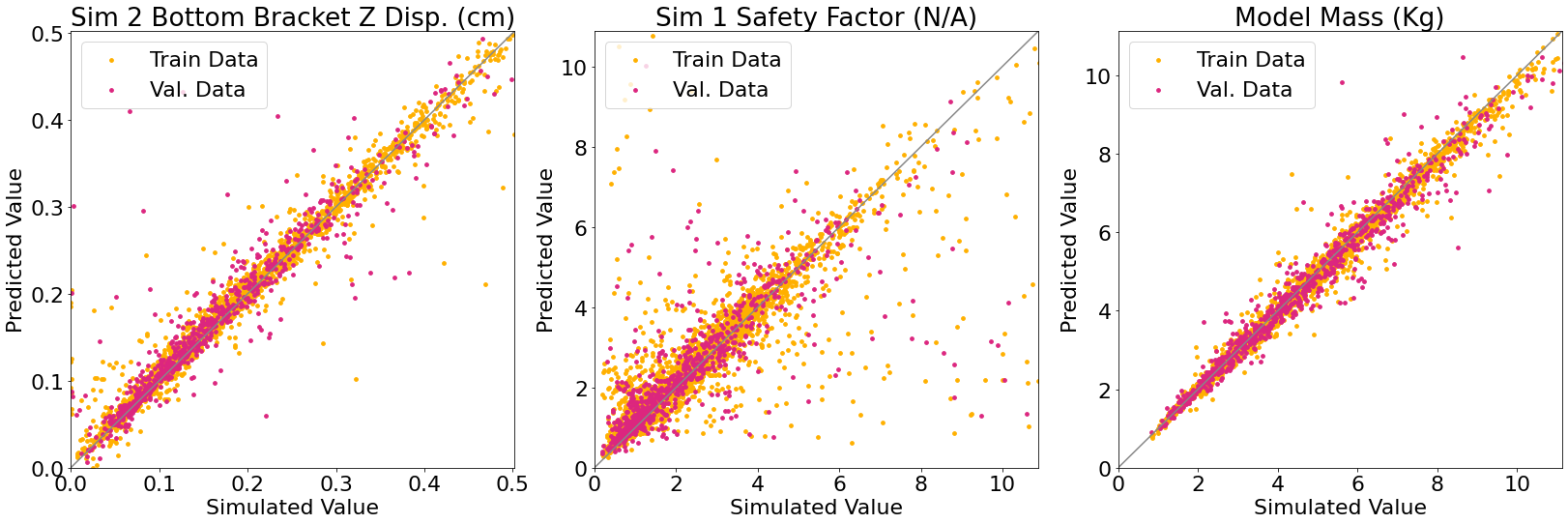}
    \caption{Regression performance of AutoML model on select performance parameters.}
    \label{fig:regression}
\end{figure*}

FRAMED provides 4118 design-performance pairs, each mapping a 37-dimensional parameter vector to a 10-dimensional performance vector. We seek to train a surrogate model on these pairs which can then be queried to predict performance vectors for new bikes. Though this task constitutes a standard supervised regression problem, selecting an optimal surrogate model and optimal training parameters for the selected model is highly dataset-dependent and constantly evolves as new methods are introduced. We apply a state-of-the-art Automated Machine-Learning (AutoML) process~\cite{erickson2020autogluon}, which methodically tests Machine Learning algorithms and training parameters to identify a high-performing surrogate. To ensure that a candidate surrogate model generalizes to datapoints from outside the dataset, surrogates are evaluated on subsets of the dataset that are withheld from the model during training, a method known as cross-validation. The highest-performing regression model was found to be a weighted ensemble of several of the individual regressors tested. This final regressor's predictive performance on several performance metrics is visualized in Figure~\ref{fig:regression} by plotting predicted performance values against the simulated (ground truth) values for both the training and validation data. All ten plots are included in the Appendix. 

\subsection{Validating the AutoML Selection using Bayesian Hyperparameter Optimization}
Though AutoML has established itself in many classic Machine Learning problems, it has rarely been applied to design problems. As such, we validate the model selected by AutoML against several supervised regression staples: Neural Networks, XGBoost~\cite{chen2016xgboost}, K-Neighbors, and Decision Trees. For each regression algorithm, we identify optimal model architecture and training parameters (hyperparameters) using Bayesian Optimization~\cite{snoek2012practical}, testing roughly 200 hyperparameter configurations. For each hyperparameter configuration, we test 50 instantiations of the model (10x 5-fold cross-validation splits), resulting in a total of approximately 10,000 training runs. We initialize our Bayesian Optimization using an initial sampling approach based on Generalized Subset Designs (GSD)~\cite{surowiec2017generalized}. Details on the hyperparameter ranges, initial sampling density, optimal hyperparameters, and training procedure are included in the Appendix. As shown in Table~\ref{tab:regression}, the Weighted Ensembles Regressor from AutoML was found to significantly outperform optimal Decision Trees, K-Neighbors, and Neural Network models identified through Bayesian Optimization, whose (optimal) parameters are documented in the appendix. The AutoML ensemble achieved a higher Coefficient of Determination ($R^2$), lower Mean Squared Error (MSE), and lower Mean Absolute Error (MAE) than Decision Trees or K-Nearest Neighbors. It also significantly outperformed XGBoost in MAE and performed very similarly in $R^2$ and MSE. 

\begin{table}[!htb]
\centering
\caption{Performance of best regressors discovered through Bayesian hyperparameter optimization compared to Weighted Ensembles Regressor from AutoML.}
\resizebox{\textwidth}{!}{%
\begin{tabular}{|l|l|l|l|l|l|}
\hline
Metric & Decision Tree & K-Nearest Neighbors & XGBoost & Neural Network & AutoML \\ \hline
Coefficient of Determination (R2) & 0.440 & 0.306 & 0.595 & 0.545 & \textbf{0.605} \\ \hline
Mean Squared Error (MSE) & 1.156 & 1.270 & \textbf{0.978} & 1.031 & 0.995 \\ \hline
Mean Absolute Error (MAE) & 0.297 & 0.412 & 0.208 & 0.263 & \textbf{0.182} \\ \hline
\end{tabular}%
}
\label{tab:regression}
\end{table}

\subsection{Predicting and Validating Geometric Validity}
We can also leverage FRAMED's data on design validity to train supervised classifiers that identify if a given bicycle design candidate is geometrically valid. While such a classifier may be less valuable to seasoned bicycle designers, it is particularly useful in training generative methods, such as Deep Generative Machine Learning Models~\cite{regenwetter2021deep}. These generative models must be evaluated for geometric validity and may conceivably be guided during training to achieve geometric validity, which would require querying a surrogate during the training loop~\cite{regenwetter2022design}. As such, we follow a similar approach to the regression problem, using AutoML to train a high-performing classifier, which achieved an F1 score of 0.915 and overall accuracy of 0.998. We then validated the classifier's performance using Bayesian Hyperparameter Tuning, just as we did for the regression models. We test 5 common classification algorithms: Neural Networks, XGBoost, K-Neighbors, Support Vector Machines, and Decision Trees. Classifiers are evaluated using F1 score, accuracy, precision, recall, and area under the curve of the Receiver Operating Characteristic (AUC). These results are included in Table~\ref{tab:classification}, while the details of the validation study are included in the Appendix. The Weighted Ensembles Classifier identified through AutoML significantly outperformed every other surrogate in every metric. 

\begin{table}[!htb]
\centering
\caption{Performance of best classifiers discovered through Bayesian hyperparameter optimization compared
to Weighted Ensembles Classifier from AutoML}
\label{tab:classification}
\resizebox{\textwidth}{!}{%
\begin{tabular}{|l|c|c|c|c|c|c|}
\hline
Metric & Decision Tree & K-Nearest Neighbors & XGBoost & Neural Network& Support Vector Machine & AutoML. \\ \hline
F1 Score & 0.750 & 0.736 & 0.722 & 0.736 & 0.620 & \textbf{0.915} \\ 
Precision & 0.900 & 0.883 & 0.867 & 0.883 & 0.889 & \textbf{0.977} \\ 
Recall & 0.643 & 0.631 & 0.619 & 0.631 & 0.476 & \textbf{0.860} \\ 
Accuracy & 0.960 & 0.958 & 0.956 & 0.958 & 0.946 & \textbf{0.991} \\ 
ROC AUC & 0.817 & 0.844 & 0.936 & 0.892 & 0.924 & \textbf{0.998} \\ \hline
\end{tabular}%
}
\end{table}

\subsection{Discussion} AutoGluon ensembles were found to significantly outperform single models in both classification and regression in almost every metric. This significant performance gap is best explained by the known performance advantages of Machine Learning ensembles in many settings. However, while ensembles are typically implemented as a collection of identical models, often through methods like bagging, AutoGluon further capitalizes on the power of ensembles by not only encompassing many different models into an ensemble but constructing a multi-layer ensemble. This effectively allows different types of models to build on each other's strengths at different levels of the prediction process, while simultaneously improving on the weaknesses of the base learners. Due to the marked performance improvements of AutoML on the FRAMED dataset, we see little reason for practitioners to use base learners unless they have special requirements, such as a need for gradient information. 

If AutoML's dominance on FRAMED is remotely indicative of its performance on other design prediction problems, it presents a huge opportunity for the design field. At the moment, surrogate predictive models in engineering design are almost always implemented using individual models instead of ensembles~\cite{alizadeh2020managing}. The marked performance gap between the AutoML ensemble models and individuals indicates that significantly better surrogate models may be possible in many design applications. Not only do AutoML models have the potential to identify significantly stronger predictive models, but they also require no tuning, making them trivial to train. We highly encourage data-driven design researchers to leverage the power of Automated Machine Learning in their future supervised prediction problems. 

\section{Limitations}
FRAMED is the first dataset that provides both parametric and performance values for a large set of community-designed bicycle frames. However, it has a few limitations, which we discuss below.
FRAMED inherits BIKED's challenges with limited diversity in certain design parameters. We attempt to mitigate this by resampling these parameters. This resampling process makes FRAMED less suitable for studies about the existing bicycle design space and more suitable for surrogate models aiming to capture a wider portion of the design space. 

FRAMED's considerably larger and more comprehensive design space expands significantly on previous data-driven studies of bicycle frame design. Nonetheless, FRAMED's design space is still far more restricted than the real-world bicycle design space. For example, the design space only considers bikes with a conventional diamond frame and excludes other bicycle frame configurations, such as bicycles with rear suspension mechanisms. It also excludes bicycle designs with non-cylindrical tubes and bicycles made from materials other than the three we support. We hope to expand FRAMED in future work to include more types of geometries.

Though we validate FRAMED's results, we acknowledge potential inaccuracy in the simulations, especially reported stresses and safety factors. Further validation against physical bicycle frames with better-known sizing and parameters would help resolve this uncertainty. We also acknowledge that our frame modeling has a few assumptions. For example, we do not model curvature at the junctions of tubes since automating the parametrization and CAD model generation of these curves and fillets would be too complex. 

\section{Future Work}
A natural extension of FRAMED would be to perform constrained optimization using the surrogate models as performance predictors to identify a structurally optimal bicycle design. Physically fabricating and testing this design would be highly insightful. 

This research also has applications outside of bicycle design in the broader community of data-driven research. One of FRAMED's core contributions is the introduction of a dataset of 4500 bicycle frame designs as well as associated structural performance values for these designs. FRAMED is therefore well-positioned to support advancements in AI-based design tools such as performance-aware generative methods. Advanced AI-based design frameworks, such as Deep Generative Models (DGMs) have shown promising initial results on a variety of design problems. FRAMED is particularly well positioned to accelerate DGM development since not only do DGMs lack quality data and benchmark problems, most current DGMs do not account for design performance at all~\cite{regenwetter2021deep}. 

Another potential extension of this work is the incorporation of multimodal data, such as 3D models, sketches, or images.  Models trained on multimodal design data typically achieve higher performance and may be more generalizable across design representations. We plan to seek out opportunities to augment the dataset with different data modalities.

\section{Ethics, Privacy, Accessibility, and Dataset Maintainance}
The primary goal of FRAMED is to provide a dataset and tools for bicycle designers and data scientists alike. To this end, we publicly release all data and code developed. We also assert that the dataset does not disclose any sensitive or personal information. Bicycle frame models cannot be traced back to an individual creator. The data was adapted from the original BIKED dataset with the consent of the authors. BIKED's authors assert similar ethics and privacy considerations in the collection of their original data~\cite{regenwetter2022biked}. 

We also anticipate that researchers using FRAMED to train Deep Generative Models may simulate generated bicycle frames using our simulation framework to measure ground truth performance for generated designs. For example, in our recent work~\cite{regenwetter2022design} we simulate 10,000 frames generated using Deep Generative Models trained on a previous version of the dataset. The performance and feasibility values for these generated frames make for excellent auxiliary regression and classification data, which we release along with corresponding trained models. As other researchers generate and simulate designs using our framework, we plan to curate any publicly released simulation results into this auxiliary dataset. 

\section{Conclusion}
This study presents a data-driven approach to bicycle frame design, analysis, and performance prediction. We develop a dataset of 4500 individually-designed bicycle frames, simulate each in three loading conditions, and extract ten performance parameters of interest. We perform several validation studies on our data, such as comparing simulation results to physical experimental results on real bicycle frames and demonstrating convergence at the selected mesh resolution. Through our analysis, we highlight general themes across bicycle designs in the dataset and study a selection of frames in greater detail. Finally, we apply state-of-the-art Automated Machine Learning (AutoML) methods to design optimal surrogate models to predict frame performance and geometric validity. We validate the quality of these surrogates by comparing them to benchmark surrogates identified through Bayesian hyperparameter optimization. We find that the AutoML models dominate in all classification metrics and most regression metrics, improving F1 score by 22\% and reducing mean absolute error by 12.5\%. Through our dataset and analysis, we aim to provide a resource for the bicycle design community, in particular, to indirectly increase accessibility to custom bicycles and positively impact bicycle ridership. We simultaneously aim to support researchers in developing data-driven design methods like AI-based Generative Models.

\section{Acknowlodgements}
We would like to acknowledge BikeCAD for the original bicycle designs, Professor Daniel Frey for his input, and MathWorks for their support on this project. 

\newpage{}

\section{Appendix}
\subsection{Regression Performance of AutoML Framework}
\begin{figure*}[!htb]
    \centering
    \includegraphics[width=.99\textwidth]{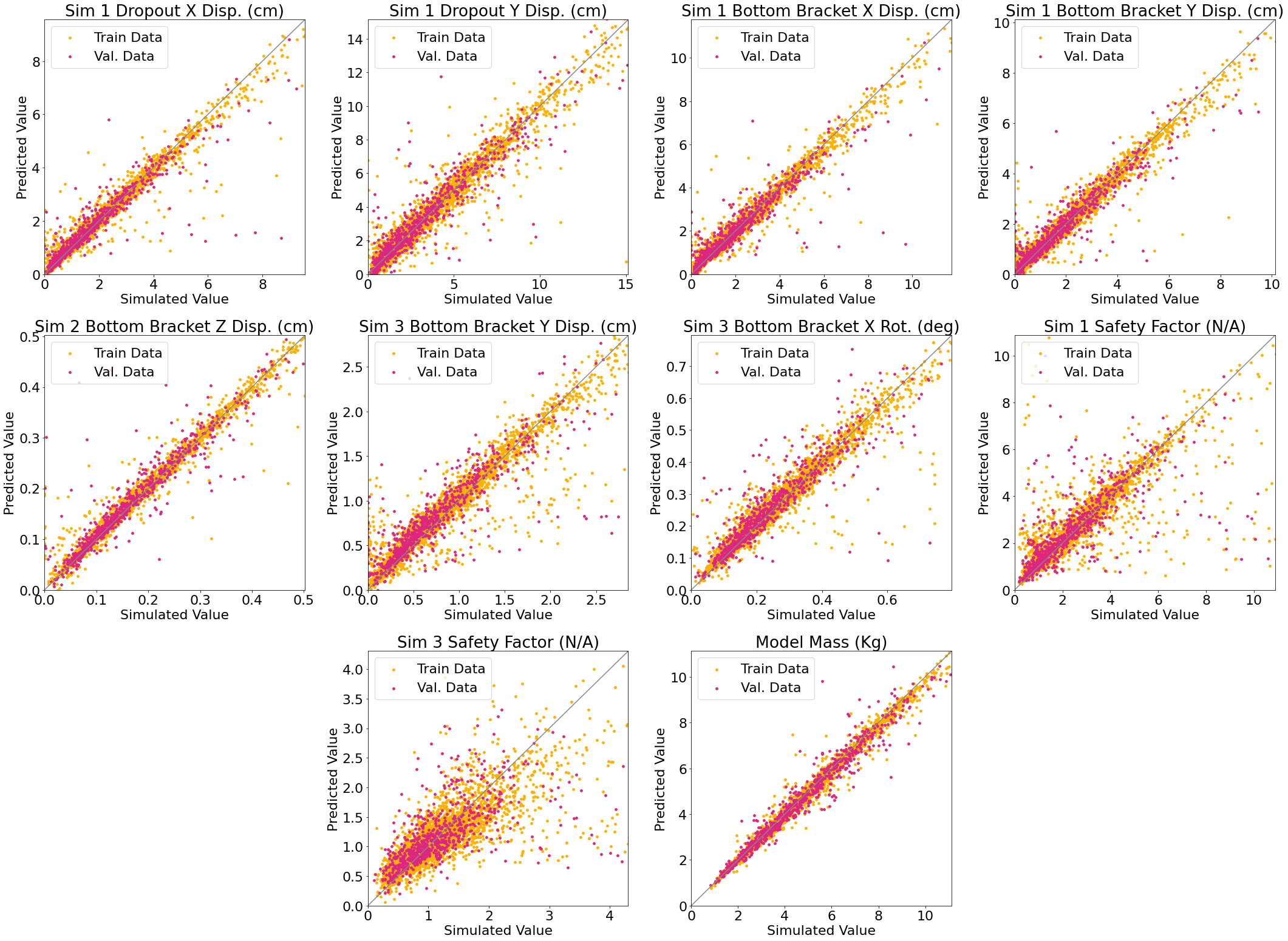}
    \caption{Regression performance of the optimal regressor selected through AutoML}
    \label{fig:regression_full}
\end{figure*}

We document the full regression performance over all ten performance objectives of the Weighted Ensembles Regressor identified through Automated Machine Learning (AutoML) in Figure~\ref{fig:regression_full}. 

\subsection{Bayesian Hyperparameter Optimization of Regression Surrogates}
This section documents the specifics of the Bayesian Hyperparameter Tuning process carried out on popular classes of regressors (Decision Trees, K-Nearest Neighbors, XGBoost, and Neural Networks). All models are initialized 50 times on ten 5-fold cross-validation splits. Bayesian Optimization is run for 100 iterations. 100 Instantiations of the optimal configuration of hyperparameters are tested to identify a top-performing individual model. All hyperparameters considered during the tuning process are listed, along with their datatypes. Ranges for integral and continuous parameters are shown and all categories are listed for categorical values. Gridpoint resolution for initial sampling is indicated in the table and Generalized Subspace Reduction factor is noted in the caption (otherwise full factorial is used). Finally, we provide the optimal value identified. 

\begin{table}[!htb]
\centering
\caption{Decision Tree Regression Hyperparameter Selection (Reduction=5)}
\label{tab:my-table}
\resizebox{\textwidth}{!}{%
\begin{tabular}{|l|c|c|c|c|c|}
\hline
Hyperparameter & Datatype & Values/{[}Min, Max{]} & Log Scaling & Gridpoint Count & Best Value \\ \hline
Maximum Tree Depth & Integer & {[}2, 10{]} & FALSE & 5 & 7 \\ \hline
Splitting Strategy & Categorical & {[}'best', 'random'{]} & FALSE & N/A & best \\ \hline
Split Quality Criterion & Categorical & {[}'squared\_error', 'friedman\_mse', 'absolute\_error'{]} & FALSE & N/A & absolute\_error \\ \hline
Min. Samples to Split Internal Node & Integer & {[}2, 20{]} & TRUE & 3 & 5 \\ \hline
Min. Samples to Split Leaf Node & Integer & {[}1, 20{]} & TRUE & 3 & 11 \\ \hline
Features Considered During Split & Integer & {[}1, 39{]} & TRUE & 3 & 32 \\ \hline
\end{tabular}%
}
\end{table}

\begin{table}[!htb]
\centering
\caption{K-Nearest-Neighbors Regression Hyperparameter Selection}
\label{tab:my-table}
\resizebox{\textwidth}{!}{%
\begin{tabular}{|l|c|c|c|c|c|}
\hline
Hyperparameter & Datatype & Values/{[}Min, Max{]} & Log Scaling & Gridpoint Count & Best Value \\ \hline
Number of Neighbors & Integer & {[}1, 1000{]} & TRUE & 10 & 70 \\ 
Neighbor Weighting & Categorical & {[}'uniform', 'distance'{]} & FALSE & N/A & distance \\ 
Distance Metric & Categorical & {[}'euclidean', 'manhattan', 'chebyshev'{]} & FALSE & N/A & euclidean \\ \hline
\end{tabular}%
}
\end{table}

\begin{table}[!htb]
\centering
\caption{XGBoost Regression Hyperparameter Selection (Reduction=5)}
\label{tab:my-table}
\resizebox{\textwidth}{!}{%
\begin{tabular}{|l|c|c|c|c|c|}
\hline
Hyperparameter & Datatype & Values/{[}Min, Max{]} & Log Scaling & Gridpoint Count & Best Value \\ \hline
Maximum Tree Depth & Integer & {[}2, 8{]} & FALSE & 3 & 6 \\ \hline
L2 Weight Regularization (lambda) & Continuous & {[}0.1, 100{]} & TRUE & 3 & 72.41212 \\ \hline
Learning Rate (eta) & Continuous & {[}0.0, 1.0{]} & FALSE & 3 & 0.275542 \\ \hline
Minimum Split Loss (gamma) & Continuous & {[}0, 1{]} & FALSE & 3 & 0.059818 \\ \hline
Minimum Child Weight & Continuous & {[}0.5, 10{]} & TRUE & 3 & 4.819113 \\ \hline
Early Stopping Patience & Integer & {[}1, 100{]} & TRUE & 3 & 8 \\ \hline
\end{tabular}%
}
\end{table}

\begin{table}[!htb]
\centering
\caption{Neural Network Regression Hyperparameter Selection (Reduction=5)}
\label{tab:my-table}
\resizebox{\textwidth}{!}{%
\begin{tabular}{|l|c|c|c|c|c|}
\hline
Hyperparameter & Datatype & Values/{[}Min, Max{]} & Log Scaling & Gridpoint Count & Best Value \\ \hline
Batch Size & Integer & {[}32, 1024{]} & TRUE & 2 & 135 \\ \hline
Early Stopping Patience & Integer & {[}1, 100{]} & TRUE & 2 & 44 \\ \hline
Dropout Rate & Continuous & {[}0.0, 1.0{]} & FALSE & 2 & 0.578647 \\ \hline
Learning Rate & Continuous & {[}1e-05, 0.1{]} & TRUE & 2 & 7.20E-05 \\ \hline
Number of Layers & Integer & {[}1, 6{]} & FALSE & 3 & 1 \\ \hline
Include Batchnormalization & Categorical & {[}False, True{]} & FALSE & N/A & FALSE \\ \hline
Activation Function & Categorical & {[}'ReLU', 'Leaky ReLU'{]} & FALSE & N/A & ReLU \\ \hline
Layer Size & Integer & {[}4, 100{]} & TRUE & 3 & 92 \\ \hline
\end{tabular}%
}
\end{table}

\subsection{Bayesian Hyperparameter Optimization of Classification Surrogates}
Bayesian hyperparameter optimization of classification surrogates is performed and documented in the same way as regression surrogates. Classifier types tested are: Decision Trees, K-Nearest Neighbors, XGBoost, Neural Networks, and Support Vector Machines. Results are documented below
\begin{table}[!htb]
\centering
\caption{Decision Tree Classification Hyperparameter Selection (Reduction=5)}
\label{tab:my-table}
\resizebox{\textwidth}{!}{%
\begin{tabular}{|l|c|c|c|c|c|}
\hline
Hyperparameter & Datatype & Values/{[}Min, Max{]} & Log Scaling & Gridpoint Count & Best Value \\ \hline
Maximum Tree Depth & Integer & {[}2, 10{]} & FALSE & 5 & 7 \\ \hline
Splitting Strategy & Categorical & {[}'best', 'random'{]} & FALSE & N/A & best \\ 
Split Quality Criterion & Categorical & {[}'gini', 'entropy'{]} & FALSE & N/A & gini \\ 
Min. Samples to Split Internal Node & Integer & {[}2, 20{]} & TRUE & 3 & 9 \\ 
Min. Samples to Split Leaf Node & Integer & {[}1, 20{]} & TRUE & 3 & 3 \\ 
Features Considered During Split & Integer & {[}1, 39{]} & TRUE & 3 & 36 \\ 
Feasible Sample Weight & Categorical & {[}'balanced', None{]} & FALSE & N/A & None \\ \hline
\end{tabular}%
}
\end{table}

\begin{table}[!htb]
\centering
\caption{K-Nearest-Neighbors Classification Hyperparameter Selection}
\label{tab:my-table}
\resizebox{\textwidth}{!}{%
\begin{tabular}{|l|l|l|l|l|l|}
\hline
Hyperparameter & Datatype & Values/{[}Min, Max{]} & Log Scaling & Gridpoint Count & Best Value \\ \hline
Number of Neighbors & Integer & {[}1, 1000{]} & TRUE & 10 & 3  \\
Neighbor Weighting & Categorical & {[}'uniform', 'distance'{]} & FALSE & N/A & distance  \\
Distance Metric & Categorical & {[}'euclidean', 'manhattan', 'chebyshev'{]} & FALSE & N/A & euclidean \\ \hline
\end{tabular}%
}
\end{table}

\begin{table}[!htb]
\centering
\caption{XGBoost Classification Hyperparameter Selection (Reduction=5)}
\label{tab:my-table}
\resizebox{\textwidth}{!}{%
\begin{tabular}{|l|l|l|l|l|l|}
\hline
Hyperparameter & Datatype & Values/{[}Min, Max{]} & Log Scaling & Gridpoint Count & Best Value \\ \hline
Maximum Tree Depth & Integer & {[}2, 8{]} & FALSE & 3 & 2 \\ \hline
L2 Weight Regularization (lambda) & Continuous & {[}0.1, 100{]} & TRUE & 3 & 0.37363 \\ \hline
Learning Rate (eta) & Continuous & {[}0.0, 1.0{]} & FALSE & 3 & 0.688255 \\ \hline
Minimum Split Loss (gamma) & Continuous & {[}0, 1{]} & FALSE & 3 & 0.695424 \\ \hline
Minimum Child Weight & Continuous & {[}0.5, 10{]} & TRUE & 3 & 0.694833 \\ \hline
Early Stopping Patience & Integer & {[}1, 100{]} & TRUE & 3 & 33 \\ \hline
\end{tabular}%
}
\end{table}

\begin{table}[!htb]
\centering
\caption{Neural Network Classification Hyperparameter Selection (Reduction=5)}
\label{tab:my-table}
\resizebox{\textwidth}{!}{%
\begin{tabular}{|l|c|c|c|c|c|}
\hline
Hyperparameter & Datatype & Values/{[}Min, Max{]} & Log Scaling & Gridpoint Count & Best Value \\ \hline
Batch Size & Integer & {[}32, 1024{]} & TRUE & 2 & 775 \\ 
Early Stopping Patience & Integer & {[}1, 100{]} & TRUE & 2 & 100 \\ 
Dropout Rate & Continuous & {[}0.0, 1.0{]} & FALSE & 2 & 0.70054 \\ 
Learning Rate & Continuous & {[}1e-05, 1.0{]} & TRUE & 2 & 0.015856 \\ 
Number of Layers & Integer & {[}1, 6{]} & FALSE & 2 & 3 \\ 
Include Batchnormalization & Categorical & {[}False, True{]} & FALSE & N/A & TRUE \\ 
Activation Function & Categorical & {[}'ReLU', 'Leaky ReLU'{]} & FALSE & N/A & ReLU \\ 
Layer Size & Integer & {[}4, 100{]} & TRUE & 2 & 18 \\ 
Feasible Sample Weight & Continuous & {[}0.01, 100.0{]} & TRUE & 2 & 0.447665 \\ \hline
\end{tabular}%
}
\end{table}

\begin{table}[!htb]
\centering
\caption{Support Vector Machine Classification Hyperparameter Selection}
\label{tab:my-table}
\resizebox{\textwidth}{!}{%
\begin{tabular}{|l|l|l|l|l|l|}
\hline
Hyperparameter & Datatype & Values/{[}Min, Max{]} & Log Scaling & Gridpoint Count & Best Value \\ \hline
SVM Kernel & Categorical & {[}'linear', 'rbf', 'sigmoid'{]} & FALSE & N/A & rbf \\ 
Kernel Coefficient & Categorical & {[}'scale', 'auto'{]} & FALSE & N/A & auto \\ 
Class Weight & Categorical & {[}'balanced', None{]} & FALSE & N/A & None \\ \hline
\end{tabular}%
}
\end{table}

 \bibliographystyle{elsarticle-num} 
 \bibliography{cas-refs}





\end{document}